\documentclass{article}
\usepackage{ja_eng}
\usepackage{graphicx}
\usepackage[ansinew]{inputenc}
\usepackage{amsmath}
\usepackage{fancyhdr}
\title{Player Re-Identification Using Body Part Appearences}

\author{
    \begin{tabular}[t]{c}
        Mahesh Bhosale \\
        A2IL, University at Buffalo \\
        \texttt{mbhosale@buffalo.edu}
    \end{tabular}
    \quad 
    \begin{tabular}[t]{c}
        Abhishek Kumar \\
        A2IL, University at Buffalo \\
        \texttt{akumar58@buffalo.edu}
    \end{tabular}
     \begin{tabular}[t]{c}
        David Doermann \\
        A2IL, University at Buffalo \\
        \texttt{doermann@buffalo.edu}
    \end{tabular}
}

\begin{document}

\maketitle
\thispagestyle{fancy}

\begin{abstract}
We propose a neural network architecture that learns body part appearances for soccer player re-identification. Our model consists of a two-stream network (one stream for appearance map extraction and the other for body part map extraction) and a bilinear-pooling layer that generates and spatially pools the body part map. Each local feature of the body part map is obtained by a bilinear mapping of the corresponding local appearance and body part descriptors. Our novel representation yields a robust image-matching feature map, which results from combining the local similarities of the relevant body parts with the weighted appearance similarity. Our model does not require any part annotation on the SoccerNet-V3 re-identification dataset to train the network. Instead, we use a sub-network of an existing pose estimation network (OpenPose) to initialize the part substream and then train the entire network to minimize the triplet loss. The appearance stream is pre-trained on the ImageNet dataset, and the part stream is trained from scratch for the SoccerNet-V3 dataset. We demonstrate the validity of our model by showing that it outperforms state-of-the-art models such as OsNet and InceptionNet. 

\end{abstract}

\section{Introduction}

\begin{figure}
\includegraphics[width=\linewidth, height=5.5cm]{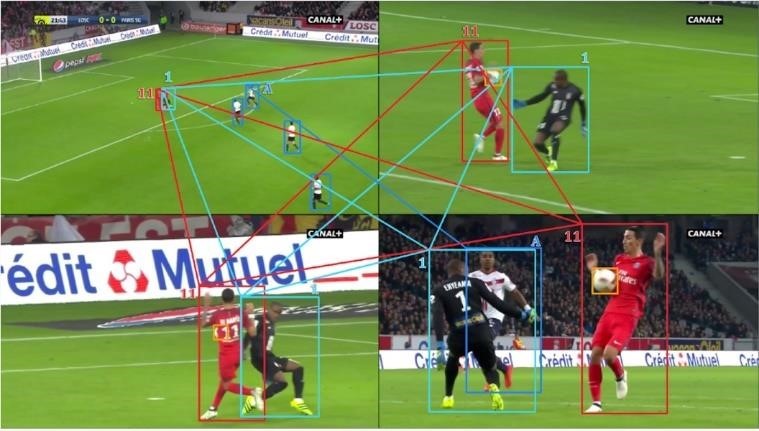}
    \caption{Association in Soccer-Player Re-identification }
    \label{fig1}
\end{figure}

The goal of person re-identification is to retrieve the images from a gallery set given an anchor image. The images in the gallery set and the anchor image are generally taken from multiple cameras from different views, making the person re-identification task challenging. Multiple camera views pose a challenge because the views are disjoint, the temporal distance between images is not constant, and lighting conditions and backgrounds are different. The challenge of person re-identification for players in any sports is even more challenging as the interclass distance is very small because of the high appearance similarity, which makes it hard to identify players even to the naked eye. Per-class samples are also very few, which renders it even harder. Fig. \ref{fig1} shows the player identity association between multiple camera views with varied image sizes and backgrounds.  

Person re-identification is necessary in many applications of video surveillance. Player re-identification has important applications in the sports analytics industry. One of the first steps in the player tracking task is player re-identification. It is also 		 
readily being used in automatic highlight generation and video assistant referee.    
Current methods of player re-identification mainly focus on two ways - one is to get high-quality discriminative features [1, 2, 3], and the other is to define the distance metric, which can be used as a loss for learning tasks [4, 5, 6].  
 
Many methods propose the use of multi-scale features \cite{bibli7} \cite{bibli8} due to their importance in the task of re-identification. However, many of these methods do not work well because of the increased challenges in player re-identification. Due to the high similarity in appearance in players similar physical and jersey appearance features are not enough to discriminate the dissimilar players from each other. Pose features should, therefore, also be considered as discriminative cues in addition to the appearance features. In particular, we use the body part features, which are learned using a subnetwork from OpenPose, which, coupled with appearance features, can address the misalignment in body parts and contribute to positional features.
\section{Methods}
\begin{figure*}
    \centering
\includegraphics[width=\linewidth]{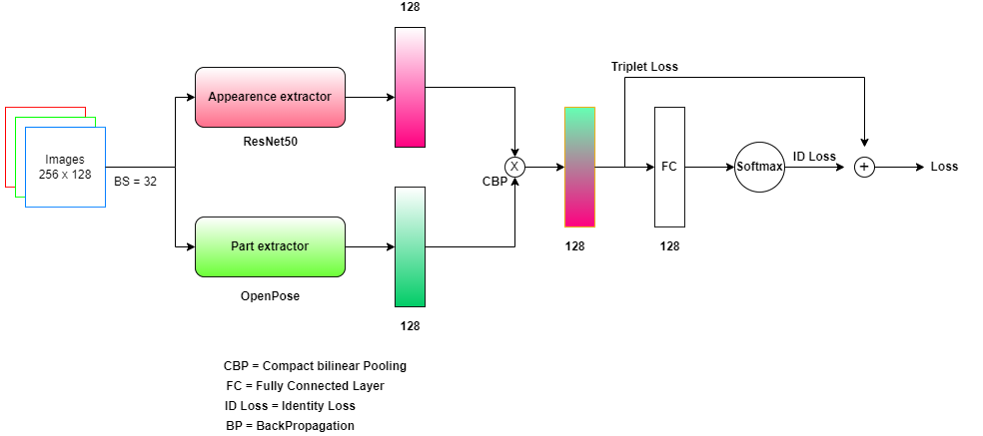}
    \caption{Architecture of proposed two-stream network
 }
    \label{fig2}
\end{figure*}

We propose to use spatial body part features along with their appearance to give us rich feature representations. We propose a two-stream network architecture; one stream works on extracting the global appearance features of the image, whilst the other stream works on extracting the body part features. Features of the two streams are combined with bilinear pooling, which allows better interaction between two features, giving a high-quality feature representation that captures the appearance of body part features. Fig. \ref{fig2} shows the architecture of the proposed model. 

\subsection{Appearence Extractor}
Appearance extractor works on extracting the appearance features of the image. We train the ResNet50 model for this task, which is initialized with pre-trained weights learned on the classification task of the Zoo dataset \cite{bibli11}. It takes the input image and outputs the appearance features $a \in H \times W$.
\subsection{Part Extractor}
Part extractor works on extracting the spatial features of body parts in an image. It takes in an input image and outputs the body part features $p \in H \times W$.
 
We do not need any annotations of body parts for this task as we train the subnetwork of OpenPose \cite{bibli10}, which is initialized with pre-trained weights on the pose estimation on the COCO dataset. 
 
OpenPose is a state-of-the-art method for pose estimation. It is a multi-stage CNN with earlier stages working on capturing the Part Affinity Fields (PAF), a 2D vector field that captures the orientation and spatial location of the body parts relative to the image domain. Later stages of CNN take PAF and produce confidence maps of 17 key points of the body parts, which are then associated using bipartite graph matching to estimate the pose.  
 
Formally, input to the first stage CNN of OpenPose is a feature map that outputs the first PAF. The first PAF with the original feature map is given as input to the next stage CNN, producing the second PAF; similarly, the outputs of the previous stages are given as input to the next stages to produce the final PAF. $Tp$ denotes the number of CNN stages producing part affinity field. 
\begin{align}
\notag
L^1 &= \phi^1(F) \\ \notag
L^t &= \phi^t(F, L^{t-1}), \forall 2\leq t\leq T_P 
\end{align}
Final PAF is given as input to the first stage CNN working on a confidence map, which outputs the first part of the confidence map. Similar to earlier stages of OpenPose working to produce PAF, the output of previous stages with the original feature map is recursively fed as input to the next stages, which finally produces the final confidence map. $Tc$ denotes the number of CNN stages producing part affinity confidence map.
\begin{align}
    \notag
    S^{T_P} &= \rho^t(F, L^{T_P}), \forall t = T_P \\ \notag
    S^{t} &= \rho^t(F, L^{T_P}, S^{t-1}), \forall T_P < t \leq T_P + T_C
\end{align}
We use the final part of the affinity confidence map St of the last CNN stage. Fig. \ref{fig4} describes the multi-stage architecture of OpenPose, with earlier stages working on PAF while later stages working on Part affinity confidence maps. Readers are strongly advised to read the OpenPose \cite{bibli10} research paper for a more detailed overview of how OpenPose works. 
\subsubsection{Bilinear Pooling}

Bilinear pooling originated as a feature extraction method for fine-grained visual recognition, where authors introduced Bilinear CNNs. It was used to classify the categories of a bird, where one stream was trained to extract the part features of a bird while the other stream was trained to extract the texture features. Similarly, we bi-pool the features from both streams, which means taking the outer product to allow finer interactions of the features. Formally, bilinear pooling is defined as, 
\begin{align}
    \notag f &= pooling_{xy}\{f_{xy}\} = \frac{1}{S}\sum_{xy}f_{xy} \\ \notag
    \text{where,} \\ \notag f_{xy} &= vec(a_{xy} \otimes p_{xy} )
\end{align}
$x$ and $y$ are spatial locations in the image and $S$ is spatial size. $a_{xy}$ is appearance feature at location ($x$, $y$) while $p_{xy}$ is part feature at location ($x$, $y$). The pooling operation we use here is average pooling. $vec(.)$ transforms a matrix to a vector, and $\otimes$ represents the outer product of two vectors, with the output being a matrix. The pooled feature, therefore, incorporates the appearance of body parts in an image and is, therefore, part-aligned. The part-aligned feature is normalized at the end,
\begin{align}
\notag
\tilde{f} = \frac{f}{\| \mathbf{f} \|_2}
\end{align}
Both the network streams can be trained end-to-end \cite{bibli12}; \ref{fig4} represents a computational graph of two streams, A and B, with loss l, showing backpropagation in progress.
\begin{figure}
    \centering
\includegraphics[width=\linewidth]{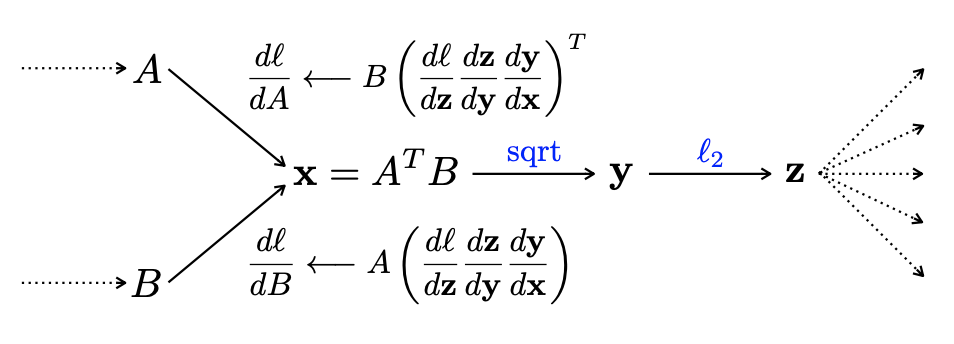}
    \caption{Back propagation in Bilinear pooling layer }
    \label{fig4}
\end{figure}

\textbf{Compact Bilinear Pooling} - To make the bilinear pooling further efficient, we use the method of compact bilinear pooling. It reduces the dimension of the vector by taking random projections to approximate the outer product, which can be computationally expensive in higher dimensions. We use a  method of the tensor sketch for projections described in \cite{bibli13}; the reader is highly advised to look at \cite{bibli13} for a detailed explanation. 

\begin{figure*}
    \centering
\includegraphics[width=\linewidth]{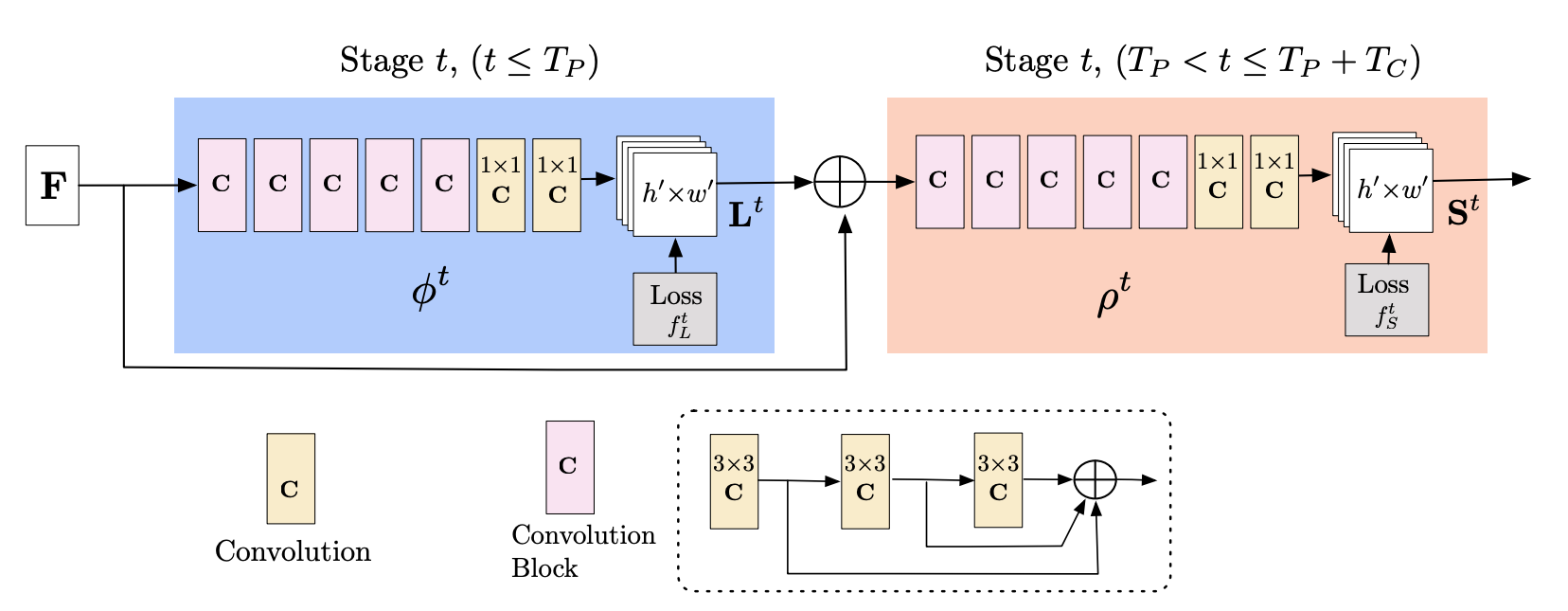}
    \caption{OpenPose Multi-Stage CNN architecture }
    \label{fig4}
\end{figure*}

\subsection{Layer Wise Similarity}
In deep CNNs, lower layers can identify some low-level features, such as shape and edges while upper layers capture the high-level features, such as semantic information of an image. Fig. \ref{fig4} shows the activation maps of multiple layers of ResNet50, starting from layer-1 at the top to layer-6 at the end.

\begin{figure}
    \centering
\includegraphics[width=\linewidth]{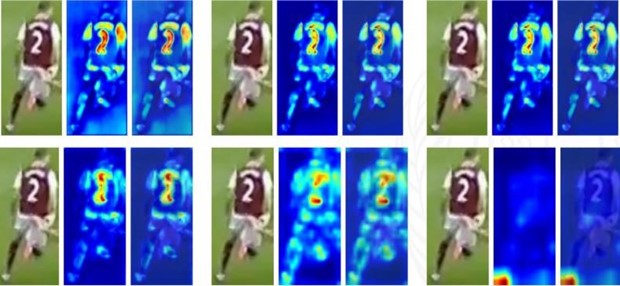}
    \caption{Activation maps ResNet50 (top-left first layer to bottom-right last layer)}
    \label{fig5}
\end{figure}

As seen from the activations maps in Fig. \ref{fig5}, activations of the last layer do not reveal much information, so we are also interested in the features extracted at some previous layers of ResNet50.

\begin{figure*}
    \centering
\includegraphics[width=\linewidth]{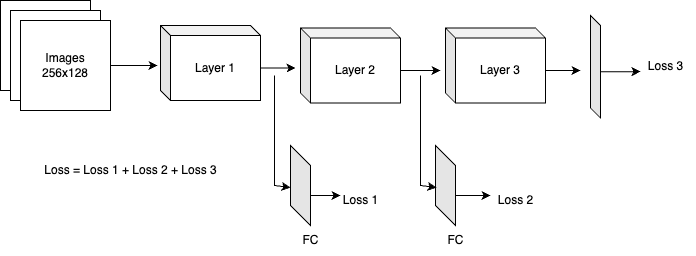}
    \caption{Proposed Layer Wise Similarity architecture }
    \label{fig6}
\end{figure*}

Fig. \ref{fig6} shows the working of the proposed Layer-wise similarity method. We take the output features of some hidden layers, add fully connected layers to enable better interactions between features, and calculate the metric learning loss on these feature vectors. We generally used similarity loss described in the next section, and since we calculate it at multiple layers, it is called layer-wise similarity loss. The number of layers, choice, and fully connected layers are tuned in cross-validation. This adds supervision, which requires feature maps of two similar images to be similar (with the use of triplet loss) in hidden layers of CNN, promoting better learning. The results section shows that adding layer-wise similarity is useful in increasing the mAP and Rank-1 scores.
\subsection{Optimization Objectives}
Person re-identification can be considered as a task of image retrieval; therefore, the image of a player under consideration is called anchor image a (taken from the query set as described in the dataset section), and the image of the same player for the same action (similar timeframe) but from other camera view (taken from the gallery set as described in dataset section) is called as positive image p whilst the image of different player for the same action (similar timeframe) from different or same view (taken from the gallery set as described in the dataset section) is called as negative image n. Triplet loss is then defined as, 
\begin{align}
\notag
    L = max(d(a, p) - d(a, n) + margin, 0)
\end{align}
where d can be any distance metric, such as L1/L2 distance. Here, we use L2 distance. Triplet loss, therefore, tries to pull two similar images (a, p) together while pushing apart the two dissimilar images (a, n). This is also generally called similarity loss. 

We can also model the problem of player re-identification as a classification task. Therefore, we can also use cross-entropy loss, which is formally defined as, 
\begin{align}
L = -\frac{1}{m}\sum_{i=1}^m y_i \cdot \log(\hat{y_i})
\end{align}
where $y_i$ is ground truth pid whilst $\hat{y_i}$ is predicted pid. It is also called identity loss. 
\section{Evaluation}

\subsection{Dataset}
\textbf{SoccerNet-V3} : We implement our model on the SoccerNet-V3 dataset. The SoccerNet-V3 Re-Identification (ReID) dataset comprises 340,993 players' thumbnails extracted from image frames of broadcast videos from 400 soccer games within 6 major leagues. Soccer players from the same team have strikingly similar appearances, making it difficult to identify them in the SoccerNet-V3 ReID dataset. Image resolution also varies significantly, and each identity has a limited number of samples, making the model more difficult to train. This makes the SoccerNet-V3 dataset more challenging. 

The SoccerNet-V3 dataset is divided into train, valid, test, and challenge sets. The training set is used for training the model. Player identification labels are formed from links between bounding boxes within an action; they can only be used within that action. As a result, player identity labels do not persist between actions, and each action where a player has been seen has a separate identity. As a result, only samples from the same action are matched against each other during the evaluation process. 

\begin{figure}
    \centering
\includegraphics[width=\linewidth]{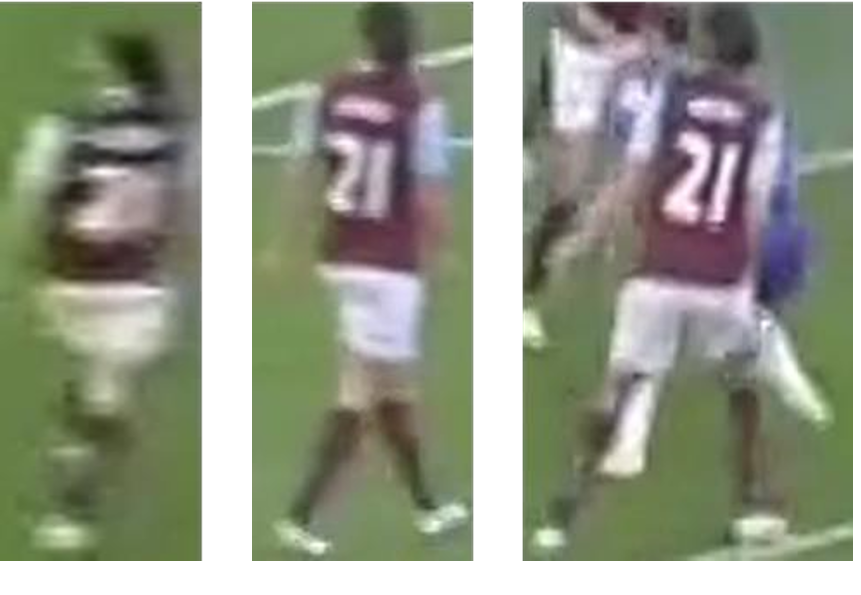}
    \caption{Same player at different actions left to right a, b, c.}
    \label{fig7}
\end{figure}

In Fig. \ref{fig7}, we can see images of a player in the same action from different viewpoints in a and b having the same pid. In c, we can use the same player in different actions with different pid. 

\subsection{Metrics}
Each valid, test, and challenge set is divided into two subsets similar to typical ReID datasets: query and gallery. The query contains images from the action frames, and the gallery contains images from the replay frames. The bounding boxes from action frames with at least one match in the replay frames are used as query samples. Bounding boxes from replay or action frames with no match make up the gallery samples. We compute a ranking of gallery samples from the same action for each query, as well as the ranking performance: rank-1 and mean average precision. As a result, query samples are only matched with gallery samples from the same action. Query samples are only matched to gallery samples from distinct camera views in traditional street surveillance ReID databases. We use  Rank-1 accuracy and mAP as evaluation metrics.

\section{Results}
In the training phase, we used a batch size of 32 and trained the model on 10\% of the dataset. We optimized our triplet loss function using Adam optimizer with a learning rate of 0. 00001. The input image resolution is kept as 256x128. For comparison, the settings and hyperparameters are kept similar for all evaluations done using different baselines on the dataset.  

\begin{table}[ht]
\begin{center}

\begin{tabular}{|c|c|c|c|}
   \hline
   \textbf{No.} & \textbf{Model} & \textbf{mAP} & \textbf{Rank} \\
\hline
1 & \textbf{Our} & \textbf{63.7} & \textbf{52.8} \\
2 & OSnet & 61.6 & 51.2 \\
3 & Inceptionv4 & 46.7 & 32 \\
4 & ResNet50\_mid & 46.5 & 31.7 \\
5 & ResNet50\_baseline & 46.7 & 32.8 \\
   \hline
\end{tabular}
\caption{Evaluation results on 10\% Soccernet data.}\label{tab1}
\end{center}
\end{table}
 
The results in tables \ref{tab1} \& \ref{tab2} show that our model can outperform the alternative baselines. In particular, we surpassed the Osnet model, which is the current state of the art for soccer player identification, by 2.1\% on mAP when training the model on a 10\% dataset. The results strongly indicate that our two-stream neural network architecture has great potential for identifying features of heterogeneous scales and viewpoints and, thus, should be considered for a broad range of visual recognition tasks. 

\begin{table}[ht]
\begin{center}

\begin{tabular}{|c|c|c|c|}
   \hline
   \textbf{No.} & \textbf{Model} & \textbf{mAP} & \textbf{Rank} \\
\hline
1 & \textbf{OSnet} & \textbf{55.5} & \textbf{45.1} \\
2 & Our & 55.0 & 42.4 \\
3 & Inceptionv4 & 49.9 & 35.8 \\
4 & ResNet50\_mid & 42.9 & 27.8 \\
5 & ResNet50\_baseline & 44.2 & 28.9 \\
   \hline
\end{tabular}
\caption{Evaluation results on 2\% Soccernet data.}\label{tab2}
\end{center}
\end{table}
 
Table \ref{tab3} shows the output of our layer-wise similarity model. We can see that by using similarity scores from intermediate levels, we were able to increase our mAP by 3.7\% and Rank-1 score by 3.6\%. The layer-wise similarity model was trained on 10\%. 

\begin{table}[ht]
\begin{center}

\begin{tabular}{|c|c|c|c|}
   \hline
   \textbf{No.} & \textbf{Model} & \textbf{mAP} & \textbf{Rank} \\
\hline
1 & \textbf{Our (layerwise)} & \textbf{50.4} & \textbf{36.2} \\
2 & ResNet & 46.7 & 32.8 \\
   \hline
\end{tabular}
\caption{Evaluation results with layerwise similarity.}\label{tab3}
\end{center}
\end{table}

\section{Conclusion}

We proposed a two-stream neural network architecture for soccer player re-identification. The key elements of our model are (1) the appearance feature, (2) the body part feature from the two-stream network, and (3) a compact bilinear pooling method to fuse the two feature maps and generate a rich body part appearance map. Extensive experiments using our model on the SoccerNet-V3 dataset demonstrated that the two-stream network could re-identify soccer players with high mean average precision and Rank-1 accuracy. We also saw that low-level features had rich semantic information, and we can further enhance our model by considering these low-level features in calculating layer-wise similarity scores and optimizing the triplet loss as future work.

\end{document}